\documentclass[lettersize,journal]{IEEEtran}
\usepackage{amsmath,amsfonts}
\usepackage{algorithmic}
\usepackage{algorithm}
\usepackage{array}
\usepackage[caption=false,font=normalsize,labelfont=sf,textfont=sf]{subfig}
\usepackage{textcomp}
\usepackage{stfloats}
\usepackage{url}
\usepackage{verbatim}
\usepackage{graphicx}
\usepackage{cite}
\usepackage{booktabs}
\begin{document}

\title{A Flow Model with Low-Rank Transformers for Incomplete Multimodal Survival Analysis}

\author{Yi~Yin, Yuntao~Shou, Zao Dai, Yun Peng, Tao~Meng, Wei~Ai, and~Keqin~Li,~\IEEEmembership{Fellow,~IEEE}
	\thanks{Corresponding Author: Tao~Meng~(mengtao@hnu.edu.cn)}
	\thanks{Y. Yin, Y. Shou, T. Meng and~W. Ai are with College of Computer and Mathematics, Central South University of Forestry and Technology, Changsha, Hunan, 410004,
		China. (isahini@csuft.edu.cn, shouyuntao@stu.xjtu.edu.cn,~mengtao@hnu.edu.cn, weiai@csuft.edu.cn)}
	\thanks{Z. Dai are with School of Computer Science and Technology, Xi'an Jiaotong University, Xi'an, Shaanxi, 710049,
	China. (daizao1102@gmail.com)}
	\thanks{K. L is with the Department of Computer Science, State University of New York, New Paltz, New York 12561, USA. (lik@newpaltz.edu)}
}

\maketitle

\begin{abstract}
	In recent years, multimodal medical data-based survival analysis has attracted much attention. However, real-world datasets often suffer from the problem of incomplete modality, where some patient modality information is missing due to acquisition limitations or system failures. Existing methods typically infer missing modalities directly from observed ones using deep neural networks, but they often ignore the distributional discrepancy across modalities, resulting in inconsistent and unreliable modality reconstruction. To address these challenges, we propose a novel framework that combines a low-rank Transformer with a flow-based generative model for robust and flexible multimodal survival prediction. Specifically, we first formulate the concerned problem as incomplete multimodal survival analysis using the multi-instance representation of whole slide images (WSIs) and genomic profiles. To realize incomplete multimodal survival analysis, we propose a class-specific flow for cross-modal distribution alignment. Under the condition of class labels, we model and transform the cross-modal distribution. By virtue of the reversible structure and accurate density modeling capabilities of the normalizing flow model, the model can effectively construct a distribution-consistent latent space of the missing modality, thereby improving the consistency between the reconstructed data and the true distribution. Finally, we design a lightweight Transformer architecture to model intra-modal dependencies while alleviating the overfitting problem in high-dimensional modality fusion by virtue of the low-rank Transformer. Extensive experiments have demonstrated that our method not only achieves state-of-the-art performance under complete modality settings, but also maintains robust and superior accuracy under the incomplete modalities scenario. 
\end{abstract}


\begin{figure}[htbp]
	\centering
	\includegraphics[width=1\linewidth]{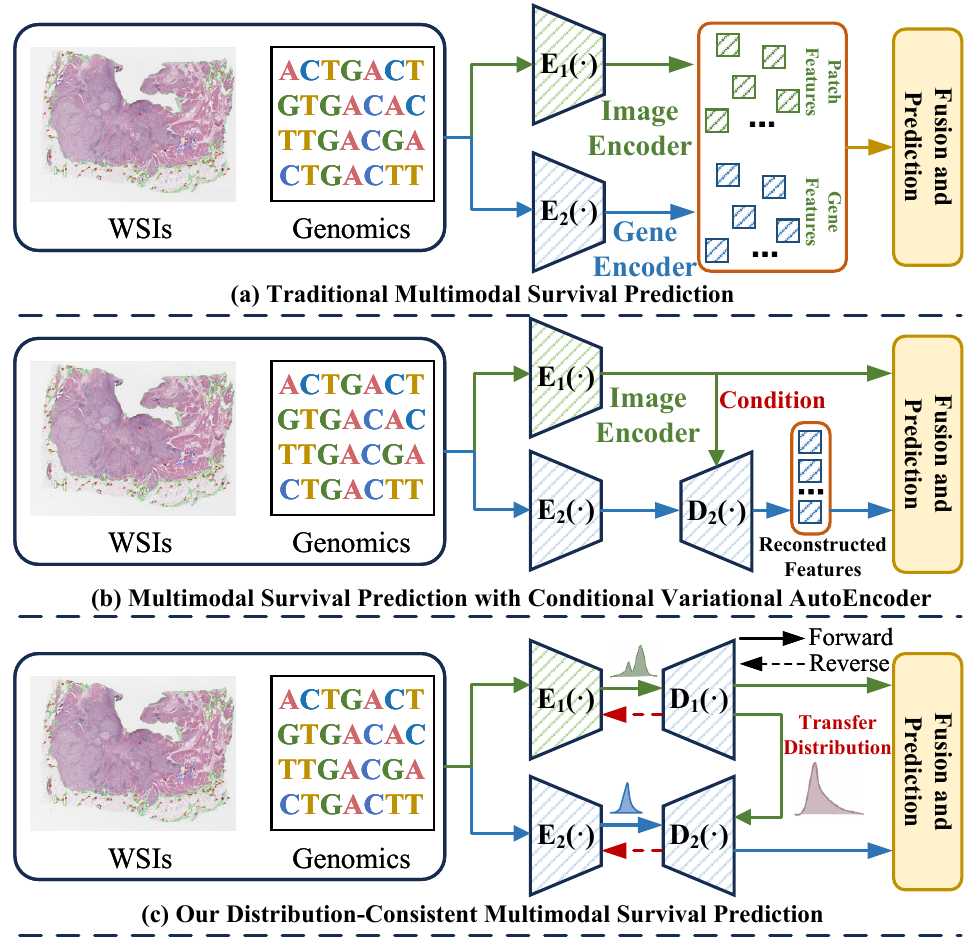}%
	\caption{Comparison of multimodal survival prediction frameworks. (a) Traditional approaches encode each modality independently and perform direct fusion for survival prediction, without explicitly modeling modality correlations. (b) Conditional VAE-based methods reconstruct missing modalities using conditional priors but may suffer from distributional inconsistency between the training and testing stages. (c) Our proposed distribution-consistent framework introduces bidirectional flow-based transformation to align modality distributions across missing and complete settings, enabling robust and consistent prediction even under modality dropout.}
	\label{fig:shuoming}
\end{figure}

\section{Introduction}

Survival analysis is a fundamental task in clinical prognosis, aiming to estimate the time until critical events such as disease progression or patient death \cite{jaume2024modeling, lv2022transsurv, shou2023comprehensive, shou2022conversational}. With the increasing availability of multimodal medical data, multimodal survival analysis has gained significant attention due to its potential to improve prediction accuracy by leveraging complementary information across different data sources \cite{jaume2024transcriptomics, liu2020supervised, shou2022object, shou2024adversarial, shou2024low}. However, in real-world clinical settings, the incomplete modality is a frequent and inevitable issue. Medical records often contain missing modalities due to device failures, acquisition costs, patient-specific constraints, or institutional variability \cite{gao2024uncertainty, tu2023deep, SHOU2025131662, shou2025masked}. This missing data severely impairs the effectiveness of multimodal fusion and poses a major challenge for building robust and generalizable survival prediction models.

As illustrated in Fig. \ref{fig:shuoming}(a), traditional frameworks independently encode each modality and fuse their embeddings for survival prediction \cite{pham2019found, zhao2021missing, meng2025gene, meng2024deep}. However, they fail to model cross-modal dependencies and cannot operate effectively when one modality is missing. To mitigate this, some approaches introduce conditional generative models such as conditional variational autoencoders, which reconstruct the missing modality from the observed one as shown in Fig. \ref{fig:shuoming}(b). While effective to some extent, these methods rely on the alignment between training-time and test-time conditional distributions, a condition often violated in real datasets, leading to degraded performance \cite{zhou2025robust, meng2024masked, shou2025spegcl}. In contrast, we propose a novel distribution-consistent framework as shown in Fig. \ref{fig:shuoming}(c), which introduces bidirectional latent alignment via flow-based transformations. 


Inspired by the above ideas, we propose a novel framework that integrates a Low-Rank Transformer with a conditional flow-based generative module for robust survival analysis under both complete and incomplete modality scenarios. By explicitly modeling the forward and backward mappings between modalities, we learn a modality-invariant representation space that is robust to both complete and missing input settings. Specifically, using the multi-instance representation of WSIs and genomic profiles, we first formulate the concerned problem as an incomplete multi-modal survival analysis. To realize incomplete multimodal survival analysis, we propose a class-specific flow for cross-modal distribution alignment. Under the condition of class labels, we model and transform the cross-modal distribution. By virtue of the reversible structure and accurate density modeling capabilities of the normalizing flow model, the model can effectively construct a distribution-consistent latent space of the missing modality, thereby improving the consistency between the reconstructed data and the true distribution. Then, we design a lightweight Transformer architecture to model intra-modal dependencies while alleviating the overfitting problem in high-dimensional modality fusion by virtue of the low-rank Transformer. The intra-modal Transformer can model the long-distance dependencies within a single modality through the self-attention mechanism: in the WSI modality, the model can capture the spatial interactions and pathological associations between different tissue regions; in the genomic profiles modality, the self-attention mechanism helps to discover potential gene co-expression structures and pathway synergy patterns, thereby improving the ability to discriminate survival outcomes. With all these designs, the final survival prediction performance is expected to be enhanced. Our method is trained in an end-to-end manner with a discrete-time survival objective, and can seamlessly handle arbitrary patterns of missing modalities during inference. Extensive experiments on public survival datasets demonstrate that our model achieves state-of-the-art performance under both fully observed and partially missing modalities, highlighting its robustness and practical applicability. Our main contributions are summarized as follows:

\begin{itemize}
	
	\item We design a class-specific flows for cross-modal distribution alignment, which can effectively construct a distribution consistency latent space of the missing modality, thereby improving the consistency between the reconstructed data and the true distribution. 
	
	\item We propose a flow model with a low-rank Transformer framework to model intra-modal and inter-modal dependencies while alleviating the overfitting problem in high-dimensional modality fusion, as well as implement cross-modal distribution transformation.
	
	
	\item We conduct extensive evaluations on multiple datasets, showing that our method outperforms existing approaches in both complete and incomplete modality settings.
\end{itemize}

\section{Related Work}

\subsection{Multimodal Survival Analysis}

Multimodal survival prediction constitutes a pivotal challenge in clinical oncology, providing clinicians with quantitative and actionable insights into disease trajectory and therapeutic efficacy \cite{vale2021long}. Historically, predictive models have predominantly leveraged structured clinical data, including short-term physiological measurements \cite{yu2021dynamic, shou2024efficient, shou2025revisiting}, longitudinal patient follow-up records \cite{capra2017assessing, shou2023graphunet, shou2025contrastive}, and radiological imaging features \cite{wang2012machine}. However, the recent proliferation of deep learning methodologies has catalyzed a paradigm shift toward histopathology-driven approaches, particularly those based on whole-slide images (WSIs). These gigapixel-scale histopathological images encapsulate rich spatial architectures and morphological phenotypes that are highly informative for prognostic modeling \cite{nakhli2023co, tran2025generating, shou2024graph, ai2025revisiting}. Given the computational intractability of processing WSIs in their native resolution, a widely adopted strategy involves decomposing each slide into a collection of smaller image patches and casting the prediction task within the multiple instance learning (MIL) framework. In this setting, slide-level survival outcomes serve as weak supervision signals shared across all constituent patches \cite{bidgoli2022evolutionary, shou2025gsdnet, shou2025multimodal}.

Concurrently, genomic profiling has become an indispensable component of modern survival analysis, enabling fine-grained risk stratification and uncovering the molecular underpinnings of tumor progression \cite{shou2024graph}. Recognizing the complementary nature of histopathological and genomic data, an increasing number of studies have sought to develop integrative models that jointly exploit these modalities to enhance both predictive performance and biological interpretability \cite{chen2021multimodal, shao2021transmil}. Notably, Zhou et al. \cite{zhouyou2024cmib} proposed CMIB, a framework that employs a co-attention mechanism to disentangle modality-specific and modality-shared representations while enforcing a multimodal information bottleneck to promote generalization. In parallel, Jaume et al. \cite{jaume2024transcriptomics} introduced TANGLE, which utilizes modality-specific encoders coupled with contrastive learning objectives to align latent embeddings across histopathological and transcriptomic views, thereby enriching slice-level semantic understanding and facilitating robust cross-modal inference.

\subsection{Incomplete Multimodal Survival Analysis}

In real world clinical applications, incomplete modality is pervasive and poses a fundamental barrier to reliable survival prediction \cite{cui2024ma}. In many care pathways, one or more data sources such as whole slide images, genomic sequencing profiles, radiology scans, or structured clinical metadata are absent for a nontrivial fraction of patients because of acquisition cost, equipment availability, privacy constraints, and heterogeneous data collection workflows \cite{gao2024uncertainty}. For example, histopathology slides are routinely archived for most cancer patients, whereas matched RNA sequencing results or comprehensive longitudinal clinical histories are often unavailable \cite{xu2025distilled}. This pattern of missingness is rarely random, is frequently tied to clinical context and resource allocation, and therefore induces distributional shifts between training and deployment cohorts that directly affect model calibration and generalization \cite{ning2021relation}.

This heterogeneity challenges multimodal survival models that typically assume complete and synchronized evidence during both training and inference. When this assumption is violated, representation learning can become biased toward the most prevalent modality, the learned cross modal relations can be underidentified, and the resulting risk estimates can be unstable and poorly calibrated. Recent lines of work attempt to improve robustness through modality dropout, shared latent representation learning, and generative imputation that leverages correlations among modalities \cite{andrew2013deep, wang2015deep, pham2019found}. These strategies can mitigate partial omissions but they commonly rely on paired multimodal supervision during training and show limited transfer to scenarios where an entire modality is systematically absent at inference \cite{pham2019found}. Furthermore, many survival analysis pipelines that center on whole slide images are designed under the premise of complete data availability, which reduces robustness to missing inputs and limits practical utility in clinical workflows where incomplete modality is the norm rather than the exception \cite{zhao2021missing}. Therefore, effectively addressing incomplete modality remains an open research problem. It demands models that can adaptively leverage available modalities, learn cross-modal correlations, and maintain reliable performance under modality-missing conditions.

\section{Problem Formulation}

Let $\mathcal{Z} = \{z_1, z_2, \dots, z_N\}$ denote a dataset of $N$ patient records, where each sample $z_i = \{x_i, c_i, y_i\}$ encompasses a whole-slide image (WSI) $x_i$, a binary censorship indicator $c_i \in \{0, 1\}$, and the observed time-to-event $y_i$. The binary censorship indicator $c_i$ serves to distinguish between two distinct types of observations: $c_i = 1$ indicates that the event of interest, typically death or disease recurrence, has been directly observed, yielding an uncensored survival time, whereas $c_i = 0$ signifies that the event remains unobserved by the end of the follow-up period, resulting in a right-censored observation. This fundamental distinction is central to survival analysis, as it acknowledges the inherent incompleteness of temporal data in clinical studies.

The primary objective of survival prediction is to estimate the discrete-time hazard function $h(y \mid x)$, which quantifies the conditional probability of an event occurring precisely at discrete time point $y$, given that the patient has survived up to and including time $y - 1$. Formally, this can be expressed as:
\begin{equation}
	h(y \mid x) = P(O = y \mid O \geq y, x), \quad y = 1, 2, \dots, K
\end{equation}
where $O$ represents the latent random variable corresponding to the true event time and $K$ denotes the total number of discrete time intervals into which the continuous follow-up period has been partitioned. This discretization allows for a more tractable computational framework while preserving the essential temporal dynamics of the survival process. The complementary survival function $S(y \mid x)$, which captures the probability that a patient survives beyond time $y$, is then defined as the cumulative product of one minus the hazard probabilities up to time $y$ as follows:
\begin{equation}
	S(y \mid x) = \prod_{j=1}^{y} \bigl(1 - h(j \mid x)\bigr)
\end{equation}
This formulation establishes a direct relationship between the instantaneous risk captured by the hazard function and the overall survival probability, providing a comprehensive probabilistic description of the patient's temporal risk profile.

We parameterize the hazard function $h(y \mid x)$ using a neural network model architecture. Specifically, we decompose the model into two primary components: a feature extractor $g(\cdot)$, which maps the high-dimensional input space of the WSI $x$ into a lower-dimensional, semantically meaningful representation, and a time-specific risk predictor $\phi_y(\cdot)$, implemented as a softmax-normalized output layer that generates the hazard probability for each discrete time interval. The resulting hazard estimate is given by:
\begin{equation}
	h(y \mid x) = \phi_y\bigl(g(x)\bigr)
\end{equation}
subject to the normalization constraint $\sum_{y=1}^K h(y \mid x) = 1$, which ensures that the predicted hazard values constitute a valid probability distribution over the discrete temporal domain. 

The model parameters are optimized through minimization of a discrete-time survival loss function, which is specifically designed to handle the presence of censored observations. This loss function, which generalizes the likelihood framework to accommodate right-censoring, is expressed as:
\begin{equation}
	\begin{split}
		\mathcal{L}_{\mathrm{surv}} = 
		& -\sum_{i=1}^N c_i \Bigl[ \log S(y_i \mid x_i) + \log h(y_i \mid x_i) \Bigr] \\
		& -\sum_{i=1}^N (1 - c_i) \log S(y_i + 1 \mid x_i)
	\end{split}
	\label{eq:l_sur}
\end{equation}
The first term in this expression penalizes the model for mispredicting the timing of observed events by jointly considering the cumulative survival probability up to time $y_i$ and the instantaneous hazard at $y_i$. The second term addresses the censored observations by penalizing the model based on the predicted survival probability beyond the censoring time $y_i$. 

%
%
%

\section{Proposed Method}

\begin{figure*}[htbp]
	\centering
	\includegraphics[width=1\linewidth]{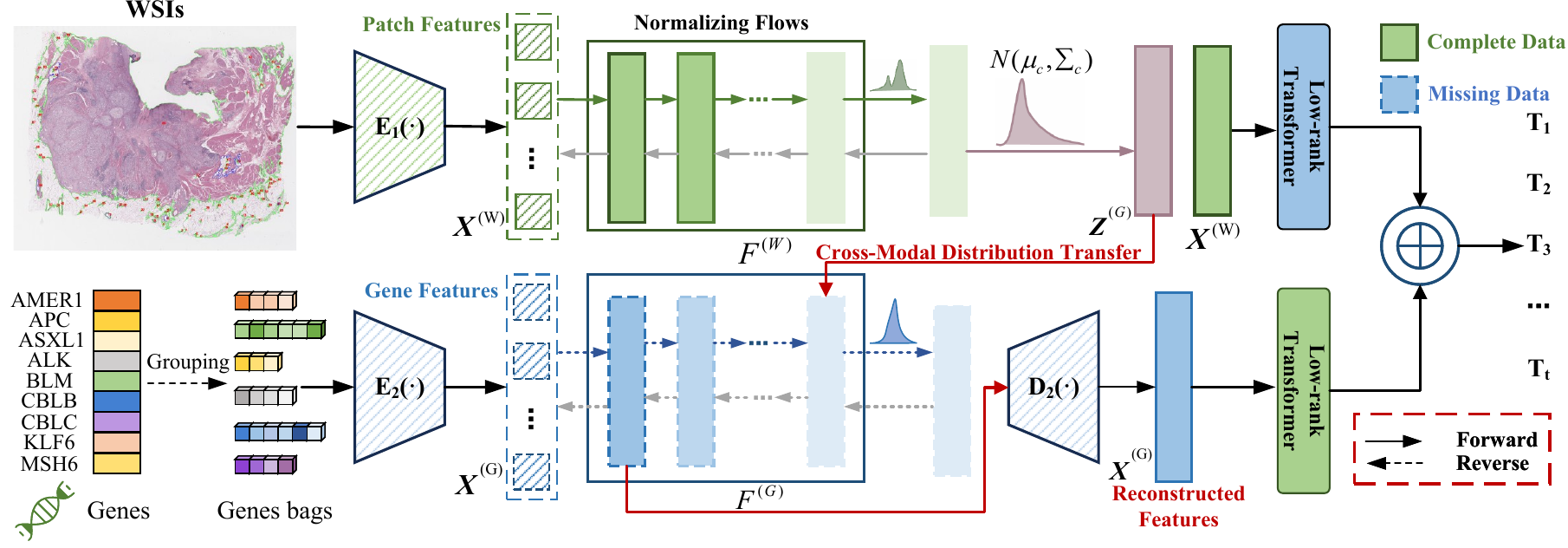}%
	\caption{ Overview of our distribution-consistent multimodal survival prediction framework. 
		Image and gene features are extracted via modality-specific encoders. 
		Visual features are modeled with normalizing flows to learn a latent distribution $\mathcal{N}(\mu_z, \Sigma_z)$, 
		while gene features are aligned via cross-modal distribution transfer and reconstructed through decoder $D(\cdot)$ if needed. 
		The fused representations are fed into a Low-rank Transformer for final survival prediction. }
	\label{fig:archi}
\end{figure*}

As illustrated in Fig.~\ref{fig:archi}, our framework takes whole-slide histopathology images and grouped genomic profiles as input and predicts patient-specific survival outcomes. The image encoder $E_1(\cdot)$ extracts patch-level visual features, which are further processed by a normalizing flow to model the underlying modality-specific latent distribution $\mathcal{N}(\mu_z, \Sigma_z)$. In parallel, the gene encoder $E_2(\cdot)$ extracts semantic representations from grouped gene embeddings. A cross-modal distribution transfer module aligns gene features with the image latent distribution, enabling robust reconstruction using decoder $D(\cdot)$ when one modality is unavailable. In this work, we take the case of missing genomic modality as a representative scenario, where only histopathology images are accessible at inference time. However, our framework is general and can be extended to other missing modality settings. The representations are then passed through a low-rank Transformer to produce the final survival predictions.

\subsection{Cross-Modal Distribution Alignment via Normalizing Flows}

Directly estimating missing modalities using deterministic mappings results in significant distribution mismatch between reconstructed and true data, ultimately degrading model performance. To mitigate this issue, we introduce a flow-based cross-modal generative module that models the conditional distribution of missing modalities given observed ones. By leveraging the expressive power of normalizing flows, we learn a reversible, distribution-aware mapping that allows the model to generate coherent representations of missing modalities.

Let $\mathbf{x}^{(o)}$ denote the observed modalities and $\mathbf{x}^{(m)}$ the missing ones for a given patient. Our goal is to model the conditional distribution $p(\mathbf{x}^{(m)} \mid \mathbf{x}^{(o)})$. We employ a conditional normalizing flow $f_\theta$ as follows:
\begin{equation}
	\mathbf{z} = f_\theta(\mathbf{x}^{(m)}; \mathbf{x}^{(o)}), \quad \mathbf{x}^{(m)} = f_\theta^{-1}(\mathbf{z}; \mathbf{x}^{(o)})
\end{equation}
where $\mathbf{z}$ follows a simple distribution (e.g., standard Gaussian), and $f_\theta$ is an invertible neural network conditioned on the observed modality embeddings. This design enables exact likelihood computation:
\begin{equation}
	\log p(\mathbf{x}^{(m)} \mid \mathbf{x}^{(o)}) = \log p(\mathbf{z}) + \log \left| \det \frac{\partial f_\theta(\mathbf{x}^{(m)}; \mathbf{x}^{(o)})}{\partial \mathbf{x}^{(m)}} \right|
\end{equation}


To maintain consistency and avoid modality collapse, we add a reconstruction loss:
\begin{equation}
	\mathcal{L}_{\text{recon}} = \mathbb{E}_{\mathbf{x}^{(m)}, \mathbf{x}^{(o)}} \left[ \| f_\theta^{-1}(f_\theta(\mathbf{x}^{(m)}; \mathbf{x}^{(o)}); \mathbf{x}^{(o)}) - \mathbf{x}^{(m)} \|_2^2 \right]
\end{equation}

To align the reconstructed features semantically with those from complete data, we further introduce a contrastive alignment loss between real and generated modality representations. Let $\mathbf{h}^{(m)}$ and $\hat{\mathbf{h}}^{(m)}$ denote the hidden representations of real and flow-generated modality features, respectively. We define the alignment loss as follows:
\begin{equation}
	\mathcal{L}_{\text{align}} =\|\hat{\mathbf{h}}^{(m)}-\mathbf{h}^{(m)}\|_{F}^{2}
\end{equation}

\subsection{Class-Specific Flows for Cross-Modal Distribution Alignment}

To improve the distributional consistency and semantic discriminability of recovered modalities, we adopt a class-specific flow strategy to model the conditional distribution transformation between observed and missing modalities. This approach mitigates the limitations of standard normalizing flows that align all modality distributions to a class-agnostic prior (e.g., $\mathcal{N}(0, I)$), which may cause latent space collapse and reduced separability among classes.

Let $\mathbf{X}^{(k)}$ denote the input of modality $k$, $k \in \{W, G\}$, and $c \in \{1, \dots, C\}$ the class label (e.g., risk level). For each modality $k$, we define a flow function $F^{(k)}$ that maps the shallow features $\mathbf{X}^{(k)} \in \mathbb{R}^{T \times d}$ into a latent space:
\begin{equation}
	\mathbf{Z}^{(k)} = F^{(k)}(\mathbf{X}^{(k)}) \sim \mathcal{N}(\boldsymbol{\mu}_c, \boldsymbol{\Sigma}_c)
\end{equation}
where $\boldsymbol{\mu}_c$ and $\boldsymbol{\Sigma}_c$ are learnable parameters defining the class-specific Gaussian distribution for class $c$. This formulation allows different classes to occupy distinct subspaces, increasing the discriminability of the latent representations.

For a missing modality $k \in \mathcal{I}_{\text{miss}}$, we estimate its latent representation $\tilde{\mathbf{Z}}^{(k)}$ by aggregating the latent features from available modalities $\mathcal{I}_{\text{obs}}$:
\begin{equation}
	\tilde{\mathbf{Z}}^{(k)} = \psi\left( \{ \mathbf{Z}^{(k)} \mid k \in \mathcal{I}_{\text{obs}} \} \right), \quad \tilde{\mathbf{Z}}^{(k)} \sim \mathcal{N}(\boldsymbol{\mu}_c, \boldsymbol{\Sigma}_c)
\end{equation}
where $\psi(\cdot)$ denotes a simple average or learned fusion function in the latent space. Then, we recover the missing modality using the inverse flow:
\begin{equation}
	\tilde{\mathbf{X}}^{(k)} = (F^{(k)})^{-1}(\tilde{\mathbf{Z}}^{(k)})
\end{equation}

Although the estimated $\widetilde{X}^{(k)}$ generally aligns with the original distribution, discrepancies from the ground truth can arise when intra-class sample dispersion is high. To address this, we introduce a lightweight decoder $\mathcal{D}^{(k)}$ to enhance the estimation, yielding the refined output $\widehat{X}^{(k)} = \mathcal{D}^{(k)}(\widetilde{X}^{(k)})$. During training, we optimize the reconstruction loss between $\widehat{X}^{(k)}$ and the corresponding ground truth $X^{(k)}$. 

To train the class-specific flows, we follow the log-likelihood principle and define the distribution-consistent loss:
\begin{equation}
	\mathcal{L}_{\text{cdt}} = - \left[ \log p_{\mathbf{Z}^{(k)}}(\mathbf{Z}^{(k)} \mid y = c) + \log \left| \det \left( \frac{\partial \mathbf{Z}^{(k)}}{\partial \mathbf{X}^{(k)}} \right) \right| \right]
\end{equation}
where $p_{\mathbf{Z}^{(k)}}(\cdot \mid y = c)$ is the density of the class-specific Gaussian, which can be explicitly written as:
\begin{equation}
	\begin{aligned}
		\log p_{\mathbf{Z}^{(k)}}(\mathbf{Z}^{(k)}) = & -\frac{d}{2} \log (2\pi) - \frac{1}{2} \log \det \boldsymbol{\Sigma}_c \\& - \frac{1}{2} (\mathbf{Z}^{(k)} - \boldsymbol{\mu}_c)^\top \boldsymbol{\Sigma}_c^{-1} (\mathbf{Z}^{(k)} - \boldsymbol{\mu}_c)
	\end{aligned}
\end{equation}

The Jacobian log-determinant is computed as the sum over affine coupling layers:
\begin{equation}
	\log \left| \det \left( \frac{\partial \mathbf{Z}^{(k)}}{\partial \mathbf{X}^{(k)}} \right) \right| = \sum_{i=1}^{L} \log \left| \det (s_i^{(k)}) \right|
\end{equation}
where $s_i^{(k)}$ denotes the scaling function in the $i$-th affine coupling layer for modality $k$.

We parameterize the class-wise Gaussian centers $\boldsymbol{\mu}_c$ and covariances $\boldsymbol{\Sigma}_c$ using zero-initialized convolutional layers:
\begin{equation}
	\boldsymbol{\mu}_c = \text{Conv}_\mu^{(c)}(\mathbf{0}), \quad \log \boldsymbol{\Sigma}_c = \text{Conv}_\Sigma^{(c)}(\mathbf{0})
\end{equation}
which allows end-to-end learning of class-specific distributions as bias terms in the convolution modules are updated.

\subsection{Low-Rank Transformer}

To effectively model complex intra-modal interactions while avoiding excessive parameter overhead, we propose a Low-Rank Multimodal Transformer (LRMT). Conventional transformer-based architectures, though powerful in capturing long-range dependencies, suffer from quadratic complexity with respect to sequence length and often result in parameter redundancy when applied to high-dimensional multimodal data. This becomes especially problematic under limited data or missing modality conditions, where overfitting is a critical concern. Inspired by recent advances in low-rank tensor modeling, we extend standard attention by introducing a low-rank bilinear decomposition of the attention score computation. Given an input feature matrix $X \in \mathbb{R}^{T \times d}$, we first compute projected query, key, and value matrices as:
\begin{equation}
	Q = X W_Q, \quad K = X W_K, \quad V = X W_V
\end{equation}
where $W_Q, W_K, W_V \in \mathbb{R}^{d \times d}$ are learnable parameters. To introduce low-rank factorization, we assume the attention weight matrix has an approximate bilinear low-rank structure:
\begin{equation}
	\text{Attn}(Q, K, V) = \text{softmax} \left( \frac{Q U A (K U)^\top}{\sqrt{d_r}} \right)(V U_V)
\end{equation}
where $U \in \mathbb{R}^{d \times d_r}$ is a shared low-rank projection across queries and keys, $A \in \mathbb{R}^{d_r \times d_r}$ is a trainable bilinear interaction matrix capturing modality-dependent mixing, and $U_V \in \mathbb{R}^{d \times d_r}$ is the value projection matrix. Alternatively, the bilinear attention kernel can be rewritten as:
\begin{equation}
	\alpha_{i,j} = \frac{(q_i^\top U) A (k_j^\top U)^\top}{\sqrt{d_r}}
\end{equation}
where $q_i$ and $k_j$ are row vectors from $Q$ and $K$, respectively. This formulation allows the model to explicitly learn structured attention patterns in a low-dimensional subspace. To further reduce redundancy, we adopt a Tucker-style decomposition where each $W_* \in \mathbb{R}^{d \times d}$ can be factorized as:
\begin{equation}
	W_* = P_* S_* R_*^\top
\end{equation}
where $P_*, R_* \in \mathbb{R}^{d \times d_r}$ and $S_* \in \mathbb{R}^{d_r \times d_r}$ are learnable matrices, ensuring compactness and expressiveness.

\subsection{Multimodal Fusion and Prediction}

The overall training objective combines survival supervision, flow-based reconstruction, and semantic alignment:
\begin{equation}
	\mathcal{L} = \mathcal{L}_{\text{surv}} + \lambda_{\text{recon}} \mathcal{L}_{\text{recon}} + \lambda_{\text{align}} \mathcal{L}_{\text{align}}
\end{equation}
where $\lambda_{\text{recon}}$ and $\lambda_{\text{align}}$ are balancing hyperparameters.

\begin{table*}[t]
	\centering
	\footnotesize
	\caption{Performance of our method across five public TCGA datasets. P and G denote the histopathological and genomic modalities, respectively. The top-performing results are indicated in \textbf{bold}, while the second-best scores are \underline{underlined}.}
	\label{tab:main} 
	\resizebox{\linewidth}{!}{%
		\begin{tabular}{lcccccccc} 
			\toprule
			\textbf{Models}         & \textbf{P} & \textbf{G} & \textbf{BLCA}          & \textbf{BRCA}          & \textbf{UCEC}          & \textbf{GBMLGG}        & \textbf{LUAD}          & \textbf{Overall} \\ 
			\midrule
			SNN~\cite{klambauer2017self}          &            & \checkmark & $0.618 \pm 0.022$      & $0.624 \pm 0.060$      & $0.679 \pm 0.040$      & $0.834 \pm 0.012$      & $0.611 \pm 0.047$      & $0.673$         \\
			SNNTrans~\cite{klambauer2017self}     &            & \checkmark & $0.645 \pm 0.042$      & $0.647 \pm 0.058$      & $0.632 \pm 0.032$      & $0.828 \pm 0.015$      & $0.633 \pm 0.049$      & $0.677$         \\
			\midrule
			AttnMIL~\cite{ilse2018attention} & \checkmark &            & $0.599 \pm 0.048$      & $0.609 \pm 0.065$      & $0.658 \pm 0.036$      & $0.818 \pm 0.025$      & $0.620 \pm 0.061$      & $0.661$         \\
			CLAM-MB~\cite{lu2021data}     & \checkmark &            & $0.565 \pm 0.027$      & $0.578 \pm 0.032$      & $0.609 \pm 0.082$      & $0.776 \pm 0.034$      & $0.582 \pm 0.072$      & $0.622$         \\
			CLAM-SB~\cite{lu2021data}     & \checkmark &            & $0.559 \pm 0.034$      & $0.573 \pm 0.044$      & $0.644 \pm 0.061$      & $0.779 \pm 0.031$      & $0.594 \pm 0.063$      & $0.629$         \\
			TransMIL~\cite{shao2021transmil}& \checkmark &            & $0.575 \pm 0.034$      & $0.666 \pm 0.029$      & $0.655 \pm 0.046$      & $0.798 \pm 0.043$      & $0.642 \pm 0.046$      & $0.667$         \\
			DeepAttnMISL~\cite{yao2020whole} & \checkmark &          & $0.504 \pm 0.042$      & $0.524 \pm 0.043$      & $0.597 \pm 0.059$      & $0.734 \pm 0.029$      & $0.548 \pm 0.050$      & $0.581$         \\
			DTFD-MIL~\cite{zhang2022dtfd} & \checkmark &        & $0.546 \pm 0.021$      & $0.609 \pm 0.059$      & $0.656 \pm 0.045$      & $0.792 \pm 0.023$      & $0.585 \pm 0.066$      & $0.638$         \\
			\midrule
			MCAT~\cite{chen2021multimodal} & \checkmark & \checkmark & $0.672 \pm 0.032$      & $0.659 \pm 0.031$      & $0.649 \pm 0.043$      & $0.835 \pm 0.024$      & $0.659 \pm 0.027$      & $0.695$         \\
			Porpoise~\cite{chen2022pan}& \checkmark & \checkmark & $0.636 \pm 0.024$      & $0.652 \pm 0.042$      & $0.695 \pm 0.032$      & $0.834 \pm 0.017$      & $0.647 \pm 0.031$      & $0.693$         \\
			MOTCat~\cite{xu2023multimodal} & \checkmark & \checkmark & $0.683 \pm 0.026$ & {$0.673 \pm 0.006$} & $0.675 \pm 0.040$ & $0.849 \pm 0.028$ & $0.670 \pm 0.038$ & $0.710$ \\
			HFBSurv~\cite{li2022hfbsurv} & \checkmark & \checkmark & $0.639 \pm 0.027$ & $0.647 \pm 0.034$ & $0.642 \pm 0.044$ & $0.838 \pm 0.013$ & $0.650 \pm 0.050$ & $0.683$ \\
			GPDBN~\cite{wang2021gpdbn} & \checkmark & \checkmark & $0.635 \pm 0.025$ & $0.654 \pm 0.033$ & $0.683 \pm 0.052$ & {$0.854 \pm 0.024$} & $0.640 \pm 0.047$ & $0.693$ \\
			CMTA~\cite{zhou2023cross}        & \checkmark & \checkmark & {$0.691 \pm 0.042$} & $0.667 \pm 0.043$ & {$0.697 \pm 0.040$} & $0.853 \pm 0.011$ & {$0.686 \pm 0.035$} & {$0.719$} \\
			LD-CVAE \cite{zhou2025robust} & \checkmark & \checkmark & $0.686 \pm 0.035$  & $0.680 \pm 0.030$  & $0.703 \pm 0.069$   & $0.849 \pm 0.017$   & $0.676 \pm 0.015$ & $0.719$ \\
			AdaMHF \cite{zhang2025adamhf}  & \checkmark & \checkmark & \underline{$0.708 \pm 0.027$} & \underline{$0.691 \pm 0.016$} & \underline{$0.716 \pm 0.041$} & \underline{$0.865 \pm 0.009$} & \underline{$0.706 \pm 0.024$} & \underline{0.737} \\
			Ours &  \checkmark &  \checkmark  &  $\textbf{0.727} \pm \textbf{0.052}$  &  $\textbf{0.714} \pm \textbf{0.024}$   &  $\textbf{0.733} \pm \textbf{0.050}$  & $\textbf{0.879} \pm \textbf{0.021}$   &  $\textbf{0.725} \pm \textbf{0.043}$ &   \textbf{0.756}   \\
			\bottomrule
		\end{tabular}
	}
\end{table*}

\section{Experiments}

\subsection{Datasets Used and Evaluation Metrics}

\textbf{Datasets.} To rigorously evaluate the effectiveness and generalizability of our proposed model, we conduct extensive experiments on five publicly available cancer cohorts from The Cancer Genome Atlas (TCGA), a large-scale and widely adopted repository that integrates multimodal clinical, genomic, and histopathological data from thousands of patients across 33 cancer types. We select five distinct cancer types that exhibit considerable diversity in both morphological features and molecular characteristics: Bladder Urothelial Carcinoma (BLCA) with 373 patients, Breast Invasive Carcinoma (BRCA) with 956 patients, Glioblastoma Multiforme and Lower Grade Glioma (GBMLGG) with 569 patients, Lung Adenocarcinoma (LUAD) with 453 patients, and Uterine Corpus Endometrial Carcinoma (UCEC) with 480 patients. Each cohort provides paired whole-slide images and corresponding genomic profiles, all annotated with ground-truth survival outcomes including event status and follow-up time. To ensure reliable and unbiased evaluation, we adopt a stratified five-fold cross-validation protocol on each dataset, maintaining the original proportion of censored and uncensored survival events in every fold. All experiments follow consistent preprocessing pipelines and partitioning procedures, enabling fair and direct comparisons with existing state-of-the-art methods.


\textbf{Evaluation Metrics.} We adopt the concordance index (C-index) as the primary evaluation metric to assess the performance of survival prediction models. The C-index measures the agreement between the predicted and actual rankings of survival times, providing an estimate of the model's ability to correctly order patients by risk. A higher C-index indicates better concordance between the predicted and true survival time order. Formally, the C-index is defined as:

\begin{equation}
	\text{C-index} = \frac{1}{n(n - 1)} \sum_{i=1}^{n} \sum_{j=1}^{n} \mathbb{I}(T_i < T_j)(1 - c_j)
\end{equation}
where $n$ is the total number of patients, $T_i$ and $T_j$ denote the survival times of the $i$-th and $j$-th patients, respectively, $c_j \in \{0,1\}$ is the censorship status with $c_j=1$ indicating the observation is censored, and $\mathbb{I}(\cdot)$ is the indicator function.

\begin{table*}[ht]
	\centering
	\footnotesize
	\caption{Benchmark results under complete missing modality settings across five public TCGA datasets, evaluated using the C-index. The highest scores are presented in \textbf{bold}, and the second-highest in \underline{underlined}.}
	\label{tab:app_miss}
	\setlength{\tabcolsep}{8.5pt}{
		\begin{tabular}{lccccccc}
			\toprule
			\textbf{Model} & \textbf{Missing Type} & \textbf{BLCA} & \textbf{GBMLGG} & \textbf{BRCA} & \textbf{LUAD} & \textbf{UCEC} & \textbf{Overall} \\ \midrule
			CMTA~\cite{zhou2023cross}       & Geno.            & 0.610 $\pm$ 0.023 & 0.739 $\pm$ 0.028 & {0.618 $\pm$ 0.042} & {0.598 $\pm$ 0.021} & 0.607 $\pm$ 0.023 & {0.634}\\ 
			MCAT~\cite{chen2021multimodal}         & Geno.          & 0.606 $\pm$ 0.041 & 0.735 $\pm$ 0.035 & 0.614 $\pm$ 0.040 & 0.566 $\pm$ 0.001 & {0.621 $\pm$ 0.038} & 0.628\\ 
			PORPOISE~\cite{chen2022pan}  & Geno.          & 0.523 $\pm$ 0.001 & 0.619 $\pm$ 0.001 & 0.478 $\pm$ 0.002 & 0.567 $\pm$ 0.002 & 0.602 $\pm$ 0.005 & 0.558\\ 
			MOTCat~\cite{xu2023multimodal} & Geno.  & {0.612 $\pm$ 0.015} & {0.741 $\pm$ 0.022} & 0.608 $\pm$ 0.021 & 0.571 $\pm$ 0.036 & 0.616 $\pm$ 0.036 & 0.630 \\ 
			LD-CVAE \cite{zhou2025robust} & Geno. & \underline{0.649 $\pm$ 0.040} & \underline{0.821 $\pm$ 0.021}  &  \underline{0.641 $\pm$ 0.012} & 0.628 $\pm$ 0.008   & \underline{0.681 $\pm$ 0.044} & \underline{0.684} \\
			AdaMHF \cite{zhang2025adamhf}         & Geno.            & {0.623 $\pm$ 0.022} & {0.754 $\pm$ 0.019} & {0.624 $\pm$ 0.011} & \underline{0.632 $\pm$ 0.012} & {0.633 $\pm$ 0.011} & {0.653} \\
			Ours &    Geno.  &  \textbf{0.669 $\pm$ 0.031}   &  \textbf{0.837 $\pm$ 0.014}   &   \textbf{0.657 $\pm$ 0.024 } &  \textbf{0.653 $\pm$ 0.023}   &  \textbf{0.694 $\pm$ 0.042} & \textbf{0.702} \\
			\midrule
			CMTA~\cite{zhou2023cross}       & Patho.           & 0.625 $\pm$ 0.037 & {0.837 $\pm$ 0.021} & 0.639 $\pm$ 0.012 & {0.678 $\pm$ 0.014} & 0.622 $\pm$ 0.018 & 0.680\\ 
			MCAT~\cite{chen2021multimodal}         & Patho.         & {0.660 $\pm$ 0.034} & 0.818 $\pm$ 0.040 & 0.641 $\pm$ 0.039 & 0.647 $\pm$ 0.027 & {0.650 $\pm$ 0.042} & {0.683}\\ 
			PORPOISE~\cite{chen2022pan}  & Patho.         & 0.601 $\pm$ 0.001 & 0.790 $\pm$ 0.013 & 0.615 $\pm$ 0.003 & 0.609 $\pm$ 0.215 & 0.555 $\pm$ 0.004 & 0.634\\ 
			MOTCat~\cite{xu2023multimodal} & Patho. & 0.641 $\pm$ 0.022 & 0.831 $\pm$ 0.029 & {0.657 $\pm$ 0.033} & 0.639 $\pm$ 0.032 & 0.642 $\pm$ 0.023 & 0.682\\ 
			LD-CVAE \cite{zhou2025robust} & Patho. &  0.674 $\pm$ 0.031 & 0.824 $\pm$ 0.037 & 0.659 $\pm$ 0.041 & 0.658 $\pm$ 0.030 & 0.682 $\pm$ 0.017 &  0.699  \\
			AdaMHF \cite{zhang2025adamhf}           & Patho.           & \underline{0.698 $\pm$ 0.012} & \underline{0.855 $\pm$ 0.034} & \underline{0.669 $\pm$ 0.038} & \underline{0.691 $\pm$ 0.022} & \underline{0.684 $\pm$ 0.021} & \underline{0.719} \\ 
			Ours &    Patho.  &   \textbf{0.713 $\pm$ 0.025} &  \textbf{0.864 $\pm$  0.018}   &  \textbf{0.695 $\pm$ 0.027}   &  \textbf{0.707 $\pm$ 0.019}   & \textbf{0.712 $\pm$ 0.044}  & \textbf{0.738}  \\
			\bottomrule
	\end{tabular}}
\end{table*}

\begin{table*}[htbp]
	\centering
	\footnotesize
	\caption{We report the mean and standard deviation of the C-index across five cancer datasets, comparing our method with existing approaches designed to handle missing modalities. The best and second-best results are highlighted in bold and \underline{underline}, respectively.}
	\label{tab:missing} 
	\setlength{\tabcolsep}{7pt}{
		\begin{tabular}{lccccccc}
			\toprule
			\textbf{Model} & \textbf{Missing Type} & \textbf{BLCA} & \textbf{BRCA} & \textbf{GBMLGG} & \textbf{LUAD} & \textbf{UCEC} & \textbf{Overall} \\
			\midrule
			VAE \cite{alemi2017deep}  & Geno.     & $0.622 \pm 0.010$ & $0.598 \pm 0.029$ & $0.660 \pm 0.029$ & $0.629 \pm 0.020$ & $0.805 \pm 0.032$ & $0.663$ \\
			GAN \cite{goodfellow2014generative} & Geno.     & $0.621 \pm 0.018$ & $0.621 \pm 0.027$ & $0.793 \pm 0.045$ & $0.608 \pm 0.028$ & $0.663 \pm 0.036$ & $0.661$ \\
			MVAE \cite{wu2018multimodal} & Geno.     & $0.629 \pm 0.009$ & $0.619 \pm 0.027$ & $0.661 \pm 0.024$ & $0.790 \pm 0.018$ & $0.610 \pm 0.030$ & $0.662$ \\
			SMIL \cite{ma2021smil}  & Geno.    & $0.627 \pm 0.015$ & $0.610 \pm 0.010$ & $0.807 \pm 0.012$ & $0.608 \pm 0.035$ & $0.678 \pm 0.016$ & $0.666$ \\
			ShaSpec \cite{wang2023multi} & Geno.   & $0.630 \pm 0.031$ & $0.626 \pm 0.027$ & $0.613 \pm 0.036$ & $0.672 \pm 0.037$ & $0.810 \pm 0.024$ & $0.670$ \\
			Transformer \cite{ma2022multimodal} & Geno. & $0.629 \pm 0.022$ & $0.621 \pm 0.046$ & $0.814 \pm 0.016$ & $0.610 \pm 0.020$ & $0.673 \pm 0.012$ & $0.669$ \\
			LD-CVAE \cite{zhou2025robust} & Geno.  & $\underline{0.649 \pm 0.040}$ & $\underline{0.641 \pm 0.012}$ & $\underline{0.821 \pm 0.021}$ & ${0.628 \pm 0.008}$ & $\underline{0.681 \pm 0.044}$ & $\underline{0.684}$ \\
			AdaMHF \cite{zhang2025adamhf}    & Geno. & $0.623 \pm 0.022$ & $0.624 \pm 0.011$ & $0.754 \pm 0.019$ & $\underline{0.632 \pm 0.012}$ & $0.633 \pm 0.011$ & $0.653$ \\
			Ours &  Geno. & \textbf{0.669 $\pm$ 0.031}  &  \textbf{0.657 $\pm$ 0.024} & \textbf{0.837 $\pm$ 0.014} & \textbf{0.653 $\pm$ 0.023} & \textbf{0.694 $\pm$ 0.042} & \textbf{0.702} \\
			\bottomrule
	\end{tabular}}
\end{table*}

\subsection{Baselines}


We compare the proposed method against a comprehensive suite of representative multimodal survival prediction approaches, spanning multiple architectural paradigms. \textbf{Feedforward Neural Network (FNN)-based methods} rely on simple yet effective multilayer perceptrons to model patient risk. Among them, SNN~\cite{klambauer2017self} employs self-normalizing networks to stabilize training and improve generalization in high-dimensional clinical data. \textbf{Attention-based MIL methods} leverage soft attention mechanisms to identify diagnostically informative regions within whole-slide images. AttnMIL~\cite{ilse2018attention} pioneers this direction by learning instance-level weights for slide-level prediction. CLAM~\cite{lu2021data} extends this idea with clustering-constrained attention for interpretable subtyping. DeepAttnMISL~\cite{yao2020whole} integrates multi-instance learning with survival modeling through hierarchical attention. DTFD-MIL~\cite{zhang2022dtfd} introduces a dual-stream framework that decouples feature extraction and aggregation for improved representation fidelity. MCAT~\cite{chen2021multimodal} and Porpoise~\cite{chen2022pan} incorporate genomic features via cross-modal attention to enhance biological interpretability. CMTA~\cite{zhou2023cross} further refines modality interaction through cross-modal token alignment. \textbf{Transformer-based approaches} exploit the global context modeling capability of self-attention. TransMIL~\cite{shao2021transmil} adapts the Vision Transformer architecture to histopathology, enabling long-range dependency capture across tissue patches. AdaMHF~\cite{zhang2025adamhf} introduces adaptive modality-aware hierarchical fusion to dynamically balance histological and genomic signals. MOTCat~\cite{xu2023multimodal} extends this with modality-ordered token concatenation for structured multimodal integration. \textbf{Bilinear fusion methods} explicitly model high-order interactions between modalities. HFBSurv~\cite{li2022hfbsurv} employs hierarchical bilinear pooling to capture fine-grained cross-modality correlations, while GPDBN~\cite{wang2021gpdbn} combines Gaussian processes with deep bilinear networks for uncertainty-aware survival prediction. \textbf{Variational autoencoder (VAE)-based models} learn probabilistic latent representations to handle data heterogeneity and noise. LD-CVAE~\cite{zhou2025robust} utilizes a conditional VAE framework with latent disentanglement to improve robustness in multimodal survival analysis.

To further evaluate robustness under missing genomic data, we additionally benchmark against representative methods designed for incomplete multimodal learning. These include deep generative models such as VAEs~\cite{alemi2017deep} and GANs~\cite{goodfellow2014generative}, which impute missing modalities via learned data distributions; multimodal VAEs like MVAE~\cite{wu2018multimodal}, which unify modalities in a shared latent space; survival-specific incomplete learning frameworks such as SMIL~\cite{ma2021smil} and ShaSpec~\cite{wang2023multi}, which incorporate modality dropout or spectral regularization; and robust transformer variants including the Robust Multimodal Transformer~\cite{ma2022multimodal} and AdaMHF~\cite{zhang2025adamhf}, which adaptively recalibrate feature importance in the presence of missing inputs. LD-CVAE~\cite{zhou2025robust} is also included in this group due to its explicit handling of modality incompleteness through conditional generation. This diverse set of baselines ensures a thorough and fair assessment of our method’s performance and resilience.


\subsection{Implementation Details}

Our model is implemented in Python 3.12 using the PyTorch 2.4.1 framework, and all experiments are conducted on a server equipped with two NVIDIA A800 GPUs (80GB memory). We adopt 5-fold cross-validation on each TCGA dataset to ensure robustness and reduce evaluation variance. We follow prior works \cite{zhao2021missing} and report C-index as the evaluation metric. Dataset-specific hyperparameters are carefully tuned to accommodate the heterogeneity of each cancer type. For \textbf{BLCA}, we set $\lambda_{\text{recon}} = 0.1$, learning rate $= 0.001$, batch size $= 1$, and train for 50 epochs with $\lambda_{\text{align}} = 0.05$. For \textbf{BRCA}, we use $\lambda_{\text{recon}} = 0.5$, learning rate $= 0.005$, batch size $= 1$, 20 epochs, and $\lambda_{\text{align}} = 0.05$. On \textbf{UCEC}, we adopt a lower learning rate of $0.0001$ with $\lambda_{\text{recon}} = 0.5$, batch size $= 2$, and train for 50 epochs with $\beta = 0.10$. For \textbf{LUAD}, we set $\lambda_{\text{recon}} = 0.1$, learning rate $= 0.001$, batch size $= 2$, and train for 50 epochs with a larger $\lambda_{\text{align}} = 0.20$. Lastly, for \textbf{GBMLGG}, the model uses $\lambda_{\text{recon}} = 1.0$, learning rate $= 0.001$, batch size $= 1$, 30 epochs, and $\lambda_{\text{align}} = 0.05$. For all experiments, models are optimized using the Adam optimizer with early stopping based on validation C-index.

\subsection{Comparison with State-of-the-Art (SOTA) Methods}


Table~\ref{tab:main} presents a comprehensive evaluation of our proposed method against state-of-the-art (SOTA) multimodal integration approaches across five publicly available TCGA datasets: BLCA, BRCA, UCEC, GBMLGG, and LUAD. All experiments are conducted under the complete modality setting, where both histopathological (P) and genomic (G) data are jointly available, enabling fair comparison with existing methods that leverage full multimodal inputs. Our approach consistently achieves the highest performance across all five datasets, as measured by the C-index. Notably, we achieve an overall C-index of 0.756, significantly surpassing the best-performing baseline LDC-VAE. This improvement is statistically significant across multiple datasets, as confirmed by paired t-tests ($p<0.05$). The robustness of our method is further underscored by its consistent superiority over recent deep learning frameworks, which employ sophisticated architectures including variational autoencoders, attention mechanisms, and multiple instance learning. Moreover, despite sharing identical input modalities with prior works, our method achieves substantial gains, indicating that the proposed flow-based low-rank transformer framework enables more effective extraction and fusion of heterogeneous biological signals. This performance advantage can be attributed to two key design innovations: (1) the use of normalizing flows to learn invertible, continuous representations that preserve information fidelity during modality alignment; and (2) the low-rank attention mechanism, which reduces computational complexity while enhancing interpretability and generalization. These components collectively enable our model to capture non-linear interactions between histopathological textures and genomic profiles, even in the presence of noisy or sparse annotations. The consistent improvements across diverse cancer types suggest strong cross-dataset generalization, reinforcing the clinical relevance and scalability of our approach.

\begin{figure*}[htbp]
	\centering
	\includegraphics[width=1\linewidth]{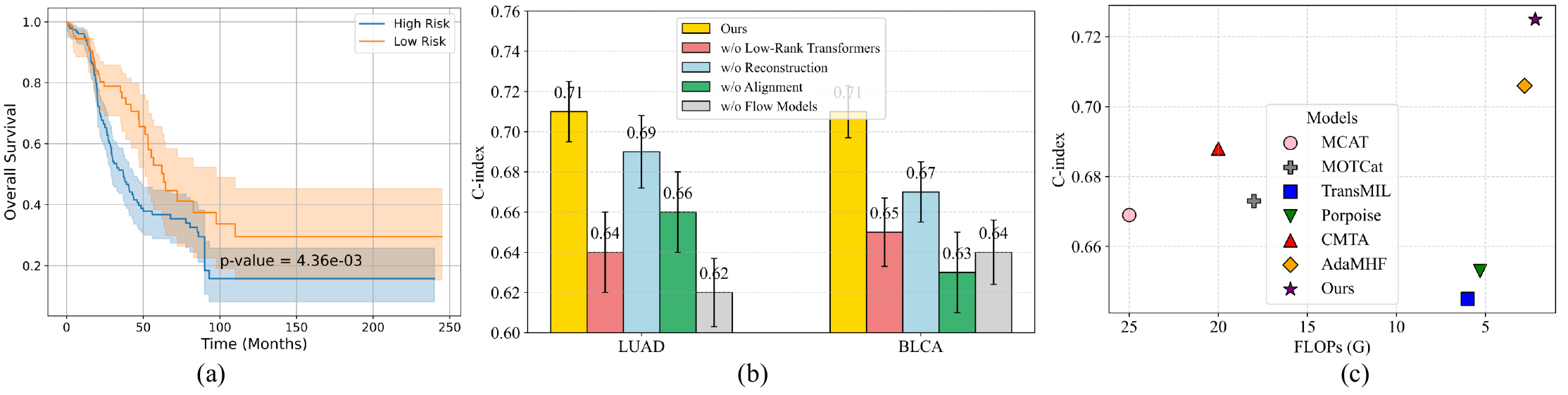}%
	\vspace{-1em}
	\caption{Quantitative and ablation analysis on the LUAD and BLCA datasets. (a) Kaplan–Meier survival curves for predicted high-risk and low-risk groups on the LUAD dataset, with a statistically significant separation. (b) Ablation study results showing the impact of removing key components on the C-index for LUAD and BLCA. (c) Comparison of computational efficiency versus predictive performance on LUAD.}
	\label{fig:three}
\end{figure*}

\subsection{Performance under Missing Modality Settings}


To further evaluate the practical robustness and generalization capability of our framework in real-world clinical settings where multimodal data are often incomplete, we conducted comprehensive assessments under scenarios where either the genomic or histopathological modality is entirely absent during inference. As illustrated in Table~\ref{tab:app_miss}, our method achieves the highest overall C-index across both missing-modality configurations, significantly outperforming all state-of-the-art baselines. In the genomic-missing setting, our model attains the best performance on all five TCGA datasets (BLCA, BRCA, UCEC, GBMLGG, LUAD) with an overall C-index of 0.702, surpassing the second-best performer LDC-VAE by 3.5\%. This consistent superiority demonstrates the model's remarkable ability to infer latent genomic signals from available histopathological features alone, leveraging the learned cross-modal correlations through its flow-based architecture.

Similarly, under the histopathological-missing scenario, our approach maintains strong predictive power, achieving the highest C-index on every dataset and an overall score of 0.738, which represents a notable improvement over the next best method AdaMHF by 4.1\%. Particularly striking gains are observed on GBMLGG and LUAD, where the absence of high-resolution histology poses significant challenges for conventional fusion models. These results underscore the effectiveness of our low-rank transformer with normalizing flows in enabling bidirectional information transfer between modalities, even when one modality is unavailable. The consistent leadership across both missing-modality conditions highlights the inherent resilience and cross-modal generalization capacity of our framework. Unlike many prior methods that rely heavily on the co-occurrence of both modalities during training and suffer severe degradation under partial input, our model exhibits robust performance due to two key design principles: the flow-based representation learning, which enables disentangled and invertible mapping between modalities allowing for reliable reconstruction of missing components; and the low-rank attention mechanism, which enhances efficiency and stability in low-data regimes by focusing on salient cross-modal interactions without overfitting.

\subsection{Compared with Baselines Addressing Missing Modality}


To further evaluate the robustness of our model in the presence of incomplete modalities, we compare its performance against several representative approaches specifically designed to handle missing genomic data. As shown in Table~\ref{tab:missing}, our method achieves the best overall performance with a mean C-index of 0.702, consistently outperforming all competing baselines across all datasets. These results demonstrate that our model effectively captures cross-modal dependencies and maintains strong predictive performance even when the genomic modality is entirely absent during inference. The consistent gains across diverse cancer types reflect the model’s ability to generalize under challenging data conditions. This robustness stems from the principled integration of normalizing flows and low-rank attention, which together enable structured representation learning and efficient cross-modal inference without relying on complete input modalities. Unlike conventional methods that degrade significantly when key data sources are missing, our framework leverages learned latent relationships to compensate for absent information, thereby preserving predictive fidelity.

\begin{figure*}[htbp]
	\centering
	\includegraphics[width=1\linewidth]{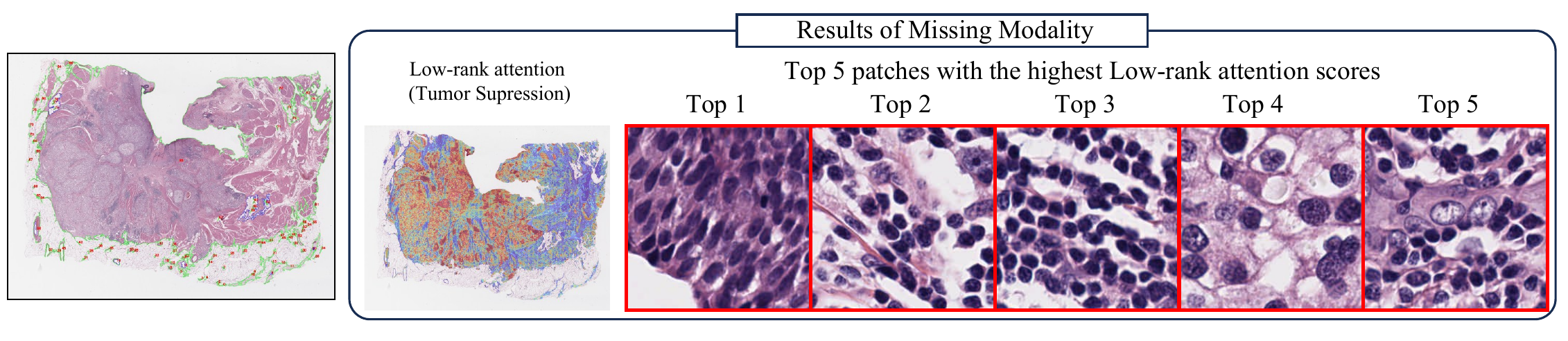}%
	\caption{Visualization of low-rank attention under the missing modality setting. From left to right: the original whole-slide image (WSI), the corresponding low-rank attention map highlighting tumor-suppressed regions, and the top-5 histopathological patches with the highest low-rank attention scores. }
	\label{fig:vis}
\end{figure*}

\subsection{Risk Stratification Analysis}


We assess the clinical utility of our model by stratifying patients from the LUAD cohort into high-risk and low-risk groups according to their predicted survival scores. This risk stratification is a standard clinical practice for prognosis and treatment planning, and its effectiveness hinges on the model’s ability to discern meaningful prognostic patterns from complex multimodal inputs. As illustrated in Fig.~\ref{fig:three}(a), the Kaplan–Meier survival curves for the two groups demonstrate a pronounced and sustained separation over the follow-up period, with patients in the low-risk group exhibiting substantially higher overall survival probabilities compared to those in the high-risk group. The statistical significance of this divergence is confirmed by the log-rank test, which yields a $p$-value of 4.36e-03. This result underscores that the risk assignments produced by our model are not only clinically interpretable but also statistically robust, reflecting genuine differences in underlying disease trajectories rather than random variation. 

\subsection{Ablation Study}


To evaluate the individual contributions of each key component in our proposed framework, we conduct a comprehensive ablation study on two representative cancer datasets, i.e., LUAD and BLCA, as illustrated in Fig. \ref{fig:three} (b). The results demonstrate that each architectural component plays a distinct and essential role in achieving robust multimodal survival prediction under incomplete data conditions. When the low-rank transformer module is removed, the model performance drops significantly across both datasets, with a C-index reduction from 0.71 to 0.64 on LUAD and from 0.71 to 0.65 on BLCA. This substantial degradation highlights the critical importance of the low-rank design in efficiently capturing cross-modal dependencies while maintaining computational and parametric efficiency. By leveraging structured weight matrices, the low-rank transformer enables effective information fusion without introducing excessive model complexity, thereby enhancing generalization under data scarcity. The removal of the reconstruction module also leads to a noticeable decline in performance, with the C-index decreasing to 0.69 on LUAD and 0.67 on BLCA. This indicates that the reconstruction objective not only regularizes the learned latent representations but also encourages the model to preserve discriminative features across modalities, particularly when certain inputs are missing during inference. Similarly, eliminating the alignment mechanism results in a drop to 0.66 on LUAD and 0.63 on BLCA, underscoring its role in enforcing consistency between modality-specific representations and facilitating reliable cross-modal integration. Notably, the most severe performance degradation is observed when the flow-based modeling component is omitted, resulting in a C-index of 0.62 on LUAD and 0.64 on BLCA. This significant drop emphasizes the crucial role of normalizing flows in accurately modeling the complex, high-dimensional joint distribution of multimodal data. By learning an invertible transformation from the observed data space to a simpler latent space, the flow model enables more precise density estimation and uncertainty-aware imputation, which is particularly beneficial in scenarios involving missing or corrupted modalities.

\begin{table}[t]
	\centering
	\caption{Comparison of computational efficiency across methods. GPU memory is measured in GB, training and test times are in seconds per epoch (for training) or per sample (for testing).}
	\label{tab:efficiency}
	\begin{tabular}{l|ccc}
		\toprule
		Methods & GPU memory & Training time & Test time \\
		\midrule
		MCAT~\cite{chen2021multimodal}     & 4.06 & 13.7 & 9.6 \\
		SurvPath~\cite{lu2021data}         & 1.95 & 8.9 & 5.3 \\
		MOTCat~\cite{xu2023multimodal}     & 3.09 & 39.2 & 38.3 \\
		CMTA~\cite{zhou2023cross}          & 19.70 & 33.1 & 28.1 \\
		AdaMHF \cite{zhang2025adamhf}      & 1.13 & 6.7 & 3.8 \\
		Ours                               & \textbf{0.45} & \textbf{3.0} & \textbf{2.2} \\
		\bottomrule
	\end{tabular}
\end{table}

\subsection{Computational Efficiency and Performance Trade-off}
%

Fig. \ref{fig:three} (c) presents a comprehensive comparison of computational efficiency and predictive performance across six recent state-of-the-art multimodal survival models on the LUAD dataset, illustrating the trade-off between model complexity and accuracy in terms of FLOPs and C-index. The results reveal a clear distinction in architectural efficiency and predictive capability among the evaluated methods. While approaches such as TransMIL and CMTA achieve moderate C-index values around 0.69 and 0.68, respectively, they incur significantly higher computational costs, with FLOPs exceeding 20 billion. Similarly, MCAT and MOTCat, though more efficient than TransMIL and CMTA, still require over 15 billion FLOPs to operate, reflecting their reliance on complex cross-modal fusion mechanisms or dense attention computations. In contrast, our proposed method achieves the highest C-index of 0.71 while maintaining the lowest computational footprint, requiring fewer than 5 billion FLOPs. This superior performance efficiency balance underscores the effectiveness of our low-rank transformer design, which reduces the parameter count and computational burden of standard self-attention mechanisms without sacrificing representational power. By factorizing the attention weight matrices into low-rank components, we enable scalable modeling of long-range dependencies across heterogeneous modalities while drastically reducing memory and computation requirements. Moreover, the positioning of our model in the lower-right region of the Pareto frontier suggests that it not only outperforms existing methods in terms of accuracy but also achieves this at a fraction of the computational cost. For instance, AdaMHF, which employs a hybrid fusion strategy and achieves a C-index of approximately 0.70, requires nearly twice the number of FLOPs compared to our model. Porpoise, despite its lightweight architecture, demonstrates limited predictive power with a C-index below 0.65, highlighting the challenge of balancing simplicity and expressiveness in multimodal learning.

Furthermore, in Table~\ref{tab:efficiency}, our proposed method demonstrates exceptional performance in terms of both memory footprint and inference speed, achieving a peak GPU memory usage of only 0.45 GB, and requiring just 3.0 seconds per epoch for training and 2.2 seconds per sample during testing. In contrast, methods such as CMTA~\cite{zhou2023cross} and MOTCat~\cite{xu2023multimodal} demand significantly higher computational resources, with GPU memory consumption exceeding 19.7 GB and 3.09 GB, respectively, due to their reliance on large-scale transformer backbones and complex cross-modal attention modules. While SurvPath~\cite{lu2021data} and AdaMHF~\cite{zhang2025adamhf} exhibit competitive training and test times, they still require more than twice the memory of our model. Notably, AdaMHF achieves fast inference (3.8 s/sample) but at the cost of moderate memory usage (1.13 GB), highlighting a trade-off between speed and resource utilization.

\subsection{Visualization Results}


To gain deeper insight into the model’s decision-making process under incomplete modality conditions, we visualize the attention distribution generated by our low-rank transformer when the genomic modality is absent during inference. As illustrated in Fig.~\ref{fig:vis}, the learned attention map effectively identifies histopathologically significant regions that are strongly associated with tumor progression and poor prognosis. The spatial distribution of attention weights highlights areas characterized by dense cellular packing, irregular nuclear morphology, and increased mitotic activity. The visualization reveals that the model focuses on biologically relevant tissue structures even in the absence of complementary genomic information. Specifically, the top five patches with the highest low-rank attention scores correspond to regions exhibiting high-grade dysplasia, abnormal nuclear pleomorphism, and disrupted tissue architecture. These findings align closely with expert pathological assessment, suggesting that the model learns clinically meaningful representations of disease severity through its attention mechanism. Notably, the attention map also demonstrates a clear suppression of non-tumorous or benign regions, indicating that the model can distinguish between malignant and normal tissue with high precision. This selective focus on tumor-relevant areas underscores the effectiveness of the low-rank attention module in filtering out irrelevant background information and enhancing signal-to-noise ratio in the feature space. By enforcing structured attention through low-rank constraints, the model avoids overfitting to spurious correlations while maintaining sensitivity to subtle but diagnostically important histological patterns.

\section{Conclusions}

In this work, we propose a novel framework that integrates a Low-Rank Transformer with a conditional flow-based generative module for robust survival analysis under both complete and incomplete modality scenarios. To realize incomplete multimodal survival analysis, we propose a class-specific flow for cross-modal distribution alignment. Under the condition of class labels, we model and transform the cross-modal distribution. By virtue of the reversible structure and accurate density modeling capabilities of the normalizing flow model, the model can effectively construct a distribution-consistent latent space of the missing modality, thereby improving the consistency between the reconstructed data and the true distribution. Then, we design a lightweight Transformer architecture to model intra-modal dependencies while alleviating the overfitting problem in high-dimensional modality fusion by virtue of the low-rank Transformer. Extensive experiments on survival datasets demonstrate that our model achieves SOTA performance under both fully observed and partially missing modalities, highlighting its robustness applicability.

\bibliographystyle{IEEEtran}
\bibliography{main}

\vfill

\end{document}